\documentclass[final,3p,times,twocolumn]{elsarticle}

\usepackage{amssymb}
\usepackage{amsmath}
\usepackage{booktabs}%
\usepackage{colortbl}
\usepackage{url}
\usepackage{subfig}
\usepackage{multirow}
\usepackage{threeparttable}

\journal{Pattern Recognition}

\begin{document}
	
	\begin{frontmatter}
		
		
		
		\title{Object-IR: Leveraging Object Consistency and Mesh Deformation for Self-Supervised Image Retargeting}
		
		
		\author[1]{Tianli Liao\corref{cor1}}
		\cortext[cor1]{Corresponding author.}
		\ead{tianli.liao@haut.edu.cn}
		\author[1]{Ran Wang}
		\ead{wran0181@163.com}
		\author[1]{Siqing Zhang}
		\ead{zsq001009@163.com}
		\author[2]{Lei Li}
		\ead{leili@haut.edu.cn}
		\author[2]{Guangen Liu}
		\ead{lgendd\_99@haut.edu.cn}
		\author[2]{Chenyang Zhao}
		\ead{zhaochy2005@163.com}
		\author[2]{Heling Cao}
		\ead{caohl@haut.edu.cn}
		\author[3]{Peng Li}
		\ead{lipeng@haut.edu.cn}
		\address[1]{Key Laboratory of Grain Information Processing and Control, Henan University of Technology, Zhengzhou, 450001, China}
		\address[2]{College of Information Science and Engineering, Henan University of Technology, Zhengzhou, 450001, China}
		\address[3]{Institute for Complexity Science, Henan University of Technology, Zhengzhou, 450001, China}
		
		\begin{abstract}
			Eliminating geometric distortion in semantically important regions remains an intractable challenge in image retargeting. This paper presents Object-IR, a self-supervised architecture that reformulates image retargeting as a learning-based mesh warping optimization problem, where the mesh deformation is guided by object appearance consistency and geometric-preserving constraints.
			Given an input image and a target aspect ratio, we initialize a uniform rigid mesh at the output resolution and 
			use a convolutional neural network to predict the motion of each mesh grid and obtain the deformed mesh.
			The retargeted result is generated by warping the input image according to the rigid mesh in the input image and the deformed mesh in the output resolution. To mitigate geometric distortion, we design a comprehensive objective function incorporating a) object-consistent loss to ensure that the important semantic objects retain their appearance, b) geometric-preserving loss to constrain simple scale transform of the important meshes, and c) boundary loss to enforce a clean rectangular output.
			Notably, our self-supervised paradigm eliminates the need for manually annotated retargeting datasets by deriving supervision directly from the input's geometric and semantic properties.
			Extensive evaluations on the RetargetMe benchmark demonstrate that our Object-IR achieves state-of-the-art performance, outperforming existing methods in quantitative metrics and subjective visual quality assessments. The framework efficiently processes arbitrary input resolutions (average inference time: 0.009s for 1024×683 resolution) while maintaining real-time performance on consumer-grade GPUs. The source code will soon be available at \url{https://github.com/tlliao/Object-IR}.
		\end{abstract}
		
		%
		
		
	\end{frontmatter}
	
		
		\section{Introduction}

		\begin{figure*}
			\centering
			\includegraphics[width=\linewidth]{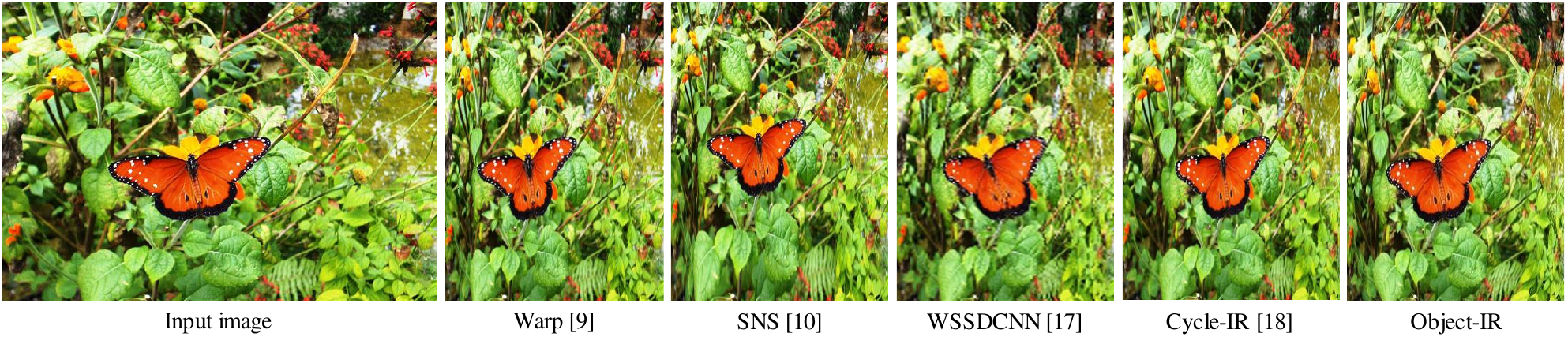}
			\caption{Image retargeting comparisons for 0.5$\times$ width resizing.}
		\label{fig:1}
	\end{figure*}
		
		Image retargeting is an advanced technique in computer vision and graphics with wide-ranging practical applications across web design, digital photography, multimedia communication, and augmented reality. Its primary goal is to adapt an image’s size or aspect ratio to suit different display platforms and devices, from large desktop screens to mobile interfaces and wearable displays. Traditional resizing methods, such as scaling and cropping, often lead to geometric distortions or the loss of critical content. Image retargeting preserves the semantic importance and spatial structure of the scene. This makes it particularly valuable in scenarios where maintaining both visual quality and contextual meaning is essential, such as delivering consistent user experiences across devices, optimizing visual content for social media and e-commerce, or ensuring clarity and interpretability in medical and aerial imaging applications.

		Various adaptive retargeting methods were proposed to ensure critical contents within images, such as objects, faces, and text, remain recognizable and visually appealing across devices. Among them, seam searching-based methods~\cite{avidan2007seam,dong2009optimized,shen2013depth,yan2014seam,zhou2016optimal,yan2019semantic,cui2020distortion,danon2021image} and image warping-based techniques~\cite{wolf2007non,wang2008optimized,guo2009image,panozzo2012robust,chang2012line,kim2018quad,patel2019reflection,cao2023polygonal} achieved good performances. The former utilizes edges, saliency, or high-level semantics to calculate the importance map and iteratively search and remove the most unnoticeable seam to change the aspect ratio. However, information loss and artifacts in salient geometrical structures may be introduced. The latter formulates retargeting as an image warping problem, in which the input image is partitioned into meshes and deformation constraints (energy terms) are imposed on the mesh grids. The optimal deformation is obtained by minimizing the corresponding energy functions. Benefiting from the high deformation flexibility of the warping model, these methods can reduce information loss and better preserve geometric structures. However, the handcrafted nature of their deformation constraints often necessitates computationally expensive iterative solvers, and they typically require trade-offs between achieving extreme retargeting ratios and maintaining geometric fidelity.
		They often fail in certain scenarios or retargeting size, as shown in Fig. \ref{fig:1}. 
		
		Recent years have witnessed growing interest in leveraging deep neural networks for image retargeting tasks~\cite{cho2017weakly,tan2019cycle,elnekave2022generating,elsner2024retargeting}. A fundamental challenge in this paradigm lies in the dataset and corresponding label construction. However, the explosion of possible aspect ratios or retargeting sizes renders conventional supervised learning approaches impractical. Current approaches circumvent this limitation through weakly- or self-supervised frameworks, where a deformation from input to output is learned by auxiliary objectives like classification consistency~\cite{cho2017weakly} or perceptual coherence~\cite{tan2019cycle}.  
		Although these approaches achieve improved geometric preservation through carefully designed objective functions, their deformation models offer limited flexibility, being restricted either to one-dimensional parameterization or to simple scaling factors.
		As illustrated in Fig. \ref{fig:1}, this architectural constraint still leads to suboptimal geometric preserving in certain retargeting scenarios.
		
		Parallel progress in geometric vision tasks using a learning-based warping framework, such as homography estimation~\cite{nie2021depth,cao2023recurrent}, image rectangling~\cite{nie2022deep}, image rectification~\cite{nie2023deep,kumari2025document}, image stitching~\cite{nie2023parallax,jia2023learning} and versatile warping model~\cite{liao2025mowa}, indicates a promising research direction. These methods constructed datasets and labels first. Then, they integrated learned warping mechanisms with geometric and pixel-level constraints, eliminating the necessity for handcrafted energy functions while maintaining geometrical structures. This framework simultaneously maintains the high-deformation freedom of warping models and the outstanding representational (or fitting) capacity of deep learning. 
		Their proven effectiveness and robustness suggest significant potential for image retargeting applications. In this paper, we demonstrate the viability of such adaptation to image retargeting tasks through a novel self-supervised neural warping framework without label construction, called \textbf{Object-IR}. Our model is trained solely by enforcing appearance consistency for important semantic objects after retargeting and constraining the meshes within these regions to undergo only scale transformations. Compared with the classification or feature consistency, the object consistency provides spatial and structural supervision, which is stronger and more generalizable.

	Specifically, we predefine a uniform rigid mesh for the output resolution and use a simple but effective convolutional neural network to estimate the grids' motion to obtain the deformed mesh. To train the network in a self-supervised way, we design a comprehensive objective function consisting of an object loss, a geometric loss, and a boundary loss. The object loss aims to enforce object consistency between the input and output images, the geometric loss is proposed to prevent the meshes within the object from distortion, and the boundary loss is included to enforce a rectangular output. To fully assess the retargeting quality, we propose a metric that uses object consistency to evaluate the distortion error in the retargeted images. Compared with the existing methods, our Object-IR can effectively mitigate the geometric distortion in the retargeted images for arbitrary retargeting sizes.
	Our contributions are summarized as follows:
	\begin{itemize}
		\item We propose a neural network that casts image retargeting into mesh-based warping and can directly produce deformed meshes given any aspect ratio.
		\item We design a comprehensive objective function that enables the retargeting method to be trained in a self-supervised way.
		\item We propose a retargeting quality metric to evaluate the distortion error in the retargeted images.
		\item Our Object-IR can be trained once and achieves the best retargeting quality given any aspect ratio.
	\end{itemize}
	
	The remainder of this paper is structured as follows. Sec. \ref{sec:related} briefly reviews prior methods and associated technical challenges. Sec. \ref{sec:method} details our proposed method. Sec. \ref{sec:implement} presents the implementation details of our Object-IR. Sec. \ref{sec:experiment} showcases the experimental results along with comparative evaluations. Finally, Sec. \ref{sec:conclude} concludes the paper.

	\begin{figure*}[t]
		\centering
		\includegraphics[width=\linewidth]{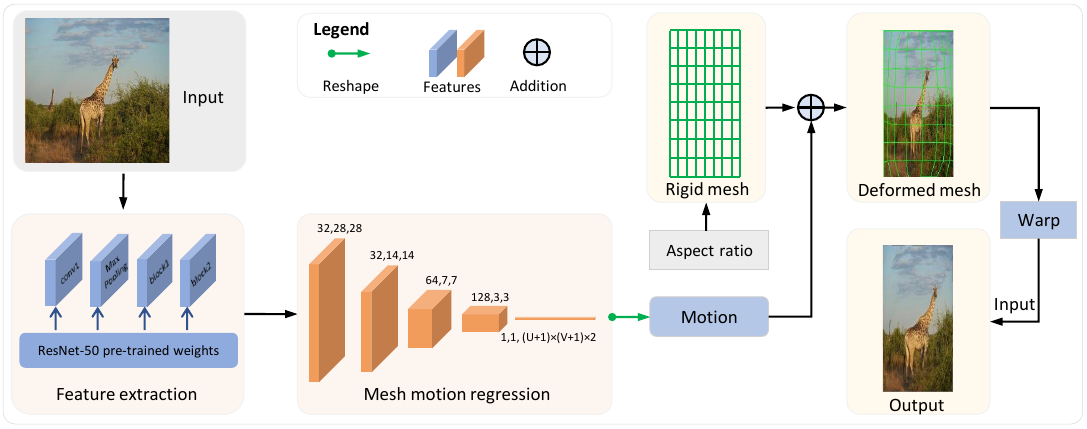}
		\caption{Overview of the proposed Object-IR. We define a rigid mesh for the output resolution and predict its motion via the regression network to obtain the deformed mesh.}
		\label{fig:framework}
	\end{figure*}

	\section{Related Work}
	\label{sec:related}
	
	This section reviews previous studies related to image retargeting, including traditional handcrafted and deep learning-based methods. For an exhaustive review, we refer the readers to surveys~\cite{pal2016content,kiess2018survey,FAN2024127416} for more details.

	\subsection{Traditional methods}
	
	Avidan and Shamir~\cite{avidan2007seam} proposed the first \emph{seam carving} (also called seam searching) operator for content-aware image retargeting. They defined an energy function via the image gradient and iteratively searched and removed unnoticeable seams to change the image to a new size with a new aspect ratio. They also supported various visual saliency measures for defining the energy. Then, different methods were proposed to enhance seam searching results by using the bidirectional image Euclidean distance~\cite{dong2009optimized}, depth map~\cite{shen2013depth}, pixel saliency~\cite{yan2014seam}, semantic segments~\cite{yan2019semantic}, and deep features~\cite{danon2021image} to define the energy function. The seam searching-based methods barely have geometric-preserving constraints and sometimes suffer from nonnegligible information loss. 
	
	Wang \emph{et al.}~\cite{wang2008optimized} proposed to formulate image retargeting as mesh-based warping and computed an optimally deformed mesh by minimizing the energy function containing the quad deformation and grid line bending energy terms. Subsequently, Panozzo \emph{et al.}~\cite{panozzo2012robust} parameterized the deformation in 1D space and efficiently solved it via a small quadratic program. They introduced as-similar and as-rigid-as-possible energy terms to prevent the image content from distortion. Chang and Chuang~\cite{chang2012line} improved the mesh deformation by preserving both salient image features and the parallelism, collinearity, and orientation properties of the line features in the images. Besides using mesh vertices' positions or distances between adjacent vertical and horizontal axes, Kim \emph{et al.}~\cite{kim2018quad} proposed to encode horizontal or vertical distance between adjacent vertices as optimization variables.
	Instead of using mesh-based warping, Dong \emph{et al.}~\cite{Dong2016Image} designed a framework based on example-based texture synthesis to enhance content-aware image retargeting. Cao \emph{et al.}~\cite{cao2023polygonal} introduced a novel method to represent deformation by high-order polygonal finite elements on a polygonal mesh with a cell distribution adapted to saliency information. It significantly extends the flexibility and capability of the deformation representation. 
	
	The warping-based methods described above can protect image structures from being distorted by introducing geometric constraints. However, the handcrafted energy terms are time-consuming to optimize and deficient in handling various shapes of objects and retargeting sizes. Besides, the energy terms often contain trade-offs between extreme retargeting size and geometric preserving. 
	
	\subsection{Deep learning-based methods}
	
	Deep learning techniques have manifested outstanding performance in various vision fields. 
	Recently, a few efforts were devoted to using learning techniques to address image retargeting.
	Cho \emph{et al.}~\cite{cho2017weakly} made the first attempt to apply deep learning to image retargeting. They introduced a weakly- and self-supervised learning framework to learn an attention map, which leads to a 1D shift map for image retargeting. The network is trained via source images and their corresponding image class annotations. Since each pixel is shifted horizontally, with similar shifts for pixels in the same column. Accordingly, its DoF equals the output image width.
	Tan \emph{et al.}~\cite{tan2019cycle} proposed a cyclic network to get rid of explicit user annotations or retargeting datasets for supervised training. They introduced a reverse mapping from the retargeted image to the input image and used a cyclic perception coherence loss for unsupervised training. They also adopt the mesh-based warping to generate a deformation field where each mesh cell is transformed via a single scale parameter. For an M$\times$N mesh, the DoF is MN.
	Elnekave and Weiss~\cite{elnekave2022generating} proposed a generative model that leverages the Sliced Wasserstein Distance to explicitly and efficiently match the distribution of patches between the input image and the generated output. The method was tested on several image generation tasks, including image retargeting. It requires no training and can generate high-quality images in a few seconds.
	Elsner \emph{et al.}~\cite{elsner2024retargeting} proposed to describe image retargeting by a displacement field that learns a similar one-dimensional deformation with~\cite{cho2017weakly} to keep the output plausible while trying to deform it only in places with low information content. It yields a more general deformation than seam carving, as it can be applied to different kinds of visual data. 
	The above methods, limited by their low deformation freedom, often struggle to produce distortion-free retargeting results. 
	In contrast, our Object-IR estimates a full 2D motion vector for each mesh grid. For an M$\times$N mesh, this yields 2MN DoF, twice that of Cycle-IR~\cite{tan2019cycle} under the same mesh resolution.
		Furthermore, unlike traditional mesh-based warping methods, our Object-IR incorporates high-dimensional latent features from the neural network, which further enlarges the representable deformation space beyond the mesh’s explicit DoF.

	\section{Proposed Method}
	\label{sec:method}

	In this section, we introduce our network structure and the objective function in Sec. \ref{sec:3-1} and Sec. \ref{sec:3-2}, respectively. Then, we introduce a metric to assess the retargeting quality in Sec. \ref{sec:3-3}. The pipeline of the proposed Object-IR is shown in Fig. \ref{fig:framework}. 
	
	\subsection{Network Structure}
	\label{sec:3-1}
	
	\subsubsection{Feature extraction}
	
	Given an input image $I$ and target aspect ratio, we adopt ResNet-50~\cite{he2016deep} with pre-trained parameters as our backbone to extract semantic features. It results in the semantic features with a resolution scaled to 1/8 of the original.
	
	\subsubsection{Mesh motion regression}
	
	We calculate a uniform rigid mesh $M_J$ for the output resolution and propose a regression module to estimate the mesh motion for $M_J$. We apply 6 convolutional layers with 3 max-pooling layers to extract deep semantic features, generating tensor features of $3\times3\times128$ size. The 3 max-pooling layers with a kernel size of 2 and a stride of 2 are applied after the 2nd, 4th, and 6th convolutional layers to progressively downsample the feature maps. Then, we flatten the tensor features into a vector of size 1152 and use a fully connected layer as the regressor to estimate the horizontal and vertical displacements of every grid vertex based on the target rigid mesh. Suppose that the rigid mesh resolution is $U\times V$, then the size of the regressor output is $(U+1)\times (V+1)\times 2$.

	\subsection{Objective Function}
	\label{sec:3-2}
	
	We optimize our network parameters using a comprehensive objective function that consists of three losses: the object loss $l_o$, the geometric loss $l_g$, and the boundary loss $l_b$. The optimization goal is formulated as follows:
	\begin{equation}
		L_{\mathrm{total}}=\lambda_o l_o+\lambda_g l_g+\lambda_b l_b
	\end{equation}
	We then give a detailed description of the three losses.
	
	\subsubsection{Object loss}
	
	\begin{figure}
		\centering
		\includegraphics[width=1\linewidth]{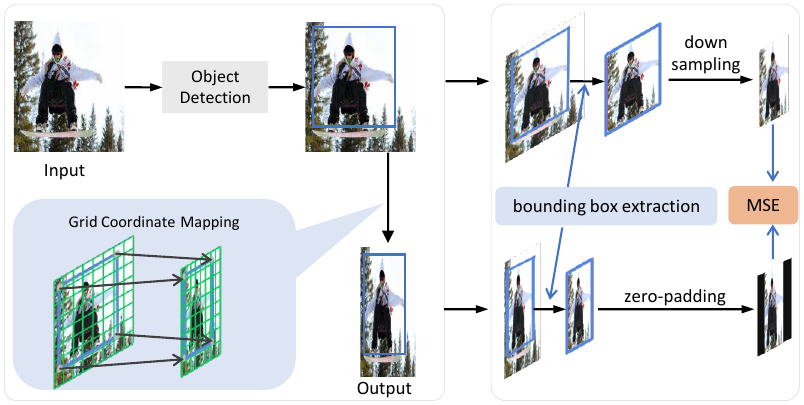}
		\caption{Illustration of the computation process for object loss.}
		\label{fig:object}
	\end{figure}
	
	Given the retargeted image $J$, we encourage object consistency with the input image $I$. Concretely, given the bounding boxes $\{O^I_i\}_{i=1}^N$ of the objects in $I$ and the bounding boxes $\{O^J_i\}_{i=1}^N$ of the corresponding objects in $J$. The object loss is defined as
	\begin{equation}
		l_o=\frac{1}{N}\sum_{i=1}^N \textit{MSE}(\mathcal{D}(O^I_i),\mathcal{P}(O^J_i)),
		\label{eq:object}
	\end{equation}
	where \textit{MSE} is the mean squared error of the two matrices. $\mathcal{D}$ and $\mathcal{P}$ denote the down-sampling and zero-padding operations, respectively. These are employed to ensure that the bounding boxes in the input and output possess the same size. Fig. \ref{fig:object} illustrates the computation process of the object loss.

	\subsubsection{Geometric loss}
	
	\begin{figure}
		\centering
		\includegraphics[width=1\linewidth]{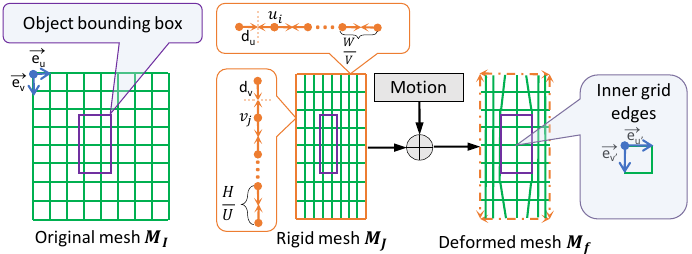}
		\caption{Illustration of the computation process for geometric loss and boundary loss.}
		\label{fig:mesh}
	\end{figure}

	To address the geometric distortion in the retargeted image, we design a geometric loss that encourages the grid edges within the objects to obey a simple scale transform, as shown in Fig. \ref{fig:mesh}. We calculate a uniform rigid mesh $M_I$ for the input image, and for each horizontal edge $\vec{e'_u}$ and vertical edge $\vec{e'_v}$ in the deformed mesh $M_f$, we calculate $l_g$ as follows:
	\begin{equation}
		\label{eq:geometric}
		l_g=\sum_{\vec{e'_u}\in M_f} \beta_{u}\left(\|\mathbf{s}\vec{e}_u-\vec{e'_u}\|\right)+\sum_{\vec{e'_v}\in M_f}\beta_{v}\left(\|\mathbf{s}\vec{e}_v-\vec{e'_v}\|\right),
	\end{equation}
	where $\vec{e}_u, \vec{e}_v$ are the corresponding edges in $M_I$, $\mathbf{s}$ is a hyper-parameter to control the scale of the transformation, defined as
	\begin{equation}
		\mathbf{s}=\sqrt{\frac{W(J)*H(J)}{W(I)*H(I)}},
		\label{eq:s}
	\end{equation}
	where $W(\cdot)$ and $H(\cdot)$ denote the width and height of an image, respectively. $\beta_{u}$ ($\beta_{v}$) decides the importance of the edge $\vec{e'_u}$ ($\vec{e'_v}$), which is computed as follows:
	\begin{equation}
		\label{eq:beta-uv}
		\beta_u (\beta_v)=\left\{\begin{array}{cc}
			1 & \vec{e}_u (\vec{e}_v) \in \cup_i O^I_i \\
			0 & \text{else}
		\end{array}\right.
	\end{equation}

	\subsubsection{Boundary loss}
	
	To generate a rectangular output image, we encourage the deformed mesh to form a rectangle as much as possible. We introduce a boundary loss to penalize the meshes as follows:
	\begin{align}
		\label{eq:boundary}
		l_b=&\sum_{u_i\in\partial_h M_J}|f_y(u_i)|+\sum_{v_j\in\partial_v M_J}|f_x(v_j)|+\nonumber\\
		&\sum_{u_i\in\partial_h M_J}\mathrm{Relu}(|f_x(u_i)|-d_u)+
		\sum_{v_j\in\partial_v M_J}\mathrm{Relu}(|f_y(v_j)|-d_v)
	\end{align}
	where $\partial_h M_J$ and $\partial_v M_J$ denote the grid vertices on the horizontal and vertical boundary of the target mesh $M_J$, respectively. $f_x$ and $f_y$ are the horizontal and vertical displacements of the grid vertices predicted by the regression module. The first two terms in Eq. (\ref{eq:boundary}) are designed to restrict the grid vertices on the horizontal (vertical) boundary such that there is no displacement in the $y$ direction ($x$ direction). The last two terms in Eq. (\ref{eq:boundary}) are included to restrict excessive displacements of grid vertices on the horizontal (vertical) boundary in the $x$ ($y$) direction\footnote{For simplicity of explanation, this paper focuses solely on width resizing. Consequently, the height of the retargeted results is identical to that of the input. The boundary loss can naturally be reformulated for height resizing.}. Fig. \ref{fig:mesh} also illustrates the computation process for boundary loss. In our experiments, $d_u$, $d_v$ are set as follows,
	\begin{equation}
		d_u=\frac{W(J)}{2V},\quad d_v=\frac{H(J)}{2U}
	\end{equation}
	
	\subsection{Retargeting Quality Assessment}
	\label{sec:3-3}
	
\begin{figure}
		\centering
		\includegraphics[width=1\linewidth]{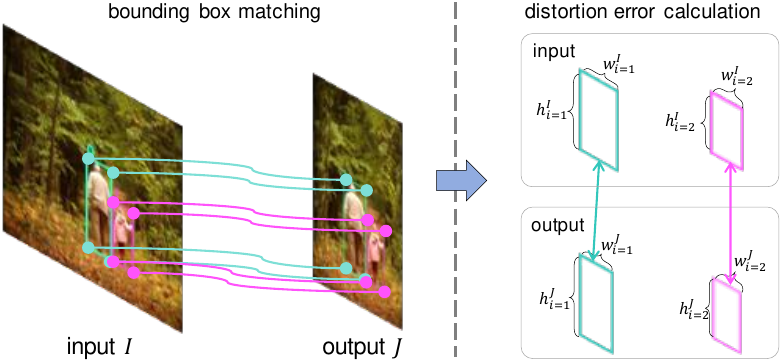}
		\caption{Illustration of the calculation of the proposed retargeting quality assessment.}
		\label{fig:quality_metric}
	\end{figure}

	We observe that, generally, no geometric distortion is introduced when the aspect ratio of the retargeting result is identical to that of the input. Geometric distortion typically occurs when the aspect ratios of the same objects in the input and output differ significantly. To evaluate the retargeting quality, we define the distortion error between the input and output as follows,
	\begin{equation}
		E(I,J)=\frac{1}{N}\sum_{i=1}^N \frac{\left|\frac{w(O^I_i)}{h(O^I_i)}-\frac{w(O^J_i)}{h(O^J_i)}\right|}{\frac{w(O^I_i)}{h(O^I_i)}},
	\end{equation}
	where $N$ is the number of objects in the input image $I$. $w(\cdot)$ and $h(\cdot)$ represent the width and height of an object's bounding box, respectively. If an object present in the input image $I$ vanishes in the output result $J$, then $w(O^J_i)/h(O^J_i)$ is set to 0. Fig. \ref{fig:quality_metric} depicts the calculation process of the distortion error. Generally, a smaller distortion error indicates higher retargeting quality. 
	
	\section{Implementation Details}
	\label{sec:implement}
	
	In this section, we present the implementation details in our Object-IR, including the training dataset and model settings.

	\subsection{Datasets}
	
	We train the proposed network using a filtered COCO dataset~\cite{lin2014microsoft}. Specifically, we select images where all instances contain detectable objects, and the area of any bounding box does not exceed half of the entire image. After filtering, a total of 18,023 images are obtained, with 17,043 used for training and 980 for testing.
	
	\subsection{Model Settings}
	
	\begin{figure}
		\centering
		\includegraphics[width=1\linewidth]{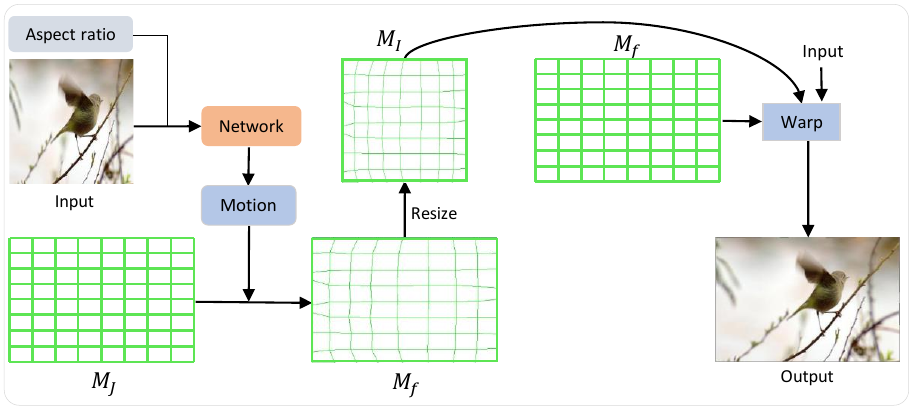}
		\caption{Retargeting process for image enlargement.}
		\label{fig:enlarge}
	\end{figure}
	
	To train our retargeting network, we resize the input images to 224$\times$224 for efficiency. Then, we use an Adam optimizer~\cite{kingma2014adam} to train our network with an exponentially decaying learning rate initialized to 1e-4. We adopt Yolo11~\cite{yolo11_ultralytics} to detect objects and extract their bounding boxes in the input images. The batch size is set to 16, and we use RELU as the activation function. 
	Input aspect ratios are randomly generated for each batch within $224/4\sim224/2$, similar to~\cite{cho2017weakly}. It takes around $3$ hours for 50 epochs on a machine with a GTX 3090 GPU.
	$\lambda_o$, $\lambda_g$, and $\lambda_b$ are set to 1, 0.1, and 0.01, respectively. $U\times V$ is set to $8\times 8$ and the implementation is based on Pytorch. We use a single 3090 GPU to finish all the training and inference.
	
	Although our model is trained solely on cases of image size reduction, our method can be directly applied to enlarge the input image. This is achieved by applying the inverse mesh deformation. Concretely, we first use the network to predict the mesh motion and calculate the deformed mesh $M_f$ for the output resolution. Then, we resize $M_f$ to the size of the input image $I$ and generate the retargeted image $J$ based on the mesh deformation from $M_f$ to $M_J$. The retargeting process for image enlargement is shown in Fig. \ref{fig:enlarge}.

	\subsection{Distortion Error Calculation}
	
	To compute the distortion error, one usually has to detect objects in the input image $I$ and the retargeted result $J$, and then establish correspondences between them. However, in our method, this step is unnecessary. The coordinates of the bounding boxes in $I$ can be directly mapped to the image space of $J$ through mesh deformation, as illustrated in Fig. \ref{fig:object} and \ref{fig:quality_metric}.

	\section{Experiments}
	\label{sec:experiment}
	
	\begin{table*}
		\centering
		\caption{Quantitative comparisons on our testing dataset.}
		\begin{tabular}{lccccc}
			\toprule
			Methods & 0.5$\times$   & 0.75$\times$  & 1.25$\times$  & 1.5$\times$ & 1.75$\times$\\
			\midrule
			CR    & 0.4522 & \textbf{0.1484} & --     & -- & --\\
			SCL   & 0.5890  & 0.3362 & 0.3075 & 0.5516 & 0.7312\\
			SC~\cite{avidan2007seam}   & 0.5751 & 0.2599 & \underline{0.2313} & 0.3621 & 0.5091\\
			SNS~\cite{wang2008optimized}  & \underline{0.3480}  & 0.2365 & 0.2726 & 0.3841 & 0.4915 \\
			WSSDCNN~\cite{cho2017weakly} & 0.3931 & 0.2877 & --     & -- & --\\
			GPDM~\cite{elnekave2022generating} & 0.5975 & 0.4495  & 0.2835  & \underline{0.3290}  &  \underline{0.4585}\\
			Object-IR & \textbf{0.3471} & \underline{0.1639} & \textbf{0.1603} & \textbf{0.3037} & \textbf{0.4382} \\
			\bottomrule
		\end{tabular}%
		\label{tab:testing}%
	\end{table*}

    \begin{table*}
		\centering
		\caption{Quantitative comparisons on RetargetMe benchmark.}
		\begin{tabular}{lccccc}
			\toprule
			Methods & 0.5$\times$   & 0.75$\times$  & 1.25$\times$  & 1.5$\times$ & 1.75$\times$\\
			\midrule
			CR    & 0.6549 & 0.4923 & --     & -- & -- \\
			SCL   & 0.5987  & 0.4161 & 0.4526 & 0.5774 & 0.7836\\
			SC~\cite{avidan2007seam}   & 0.6297 & 0.3949 & \underline{0.3709} & \underline{0.4897} & 0.7563 \\
			SNS~\cite{wang2008optimized}  & 0.4673  & \underline{0.3902}  & 0.3766 & 0.4973 & \textbf{0.5624}\\
			WSSDCNN~\cite{cho2017weakly} & 0.5050 & 0.4058 & --     & -- & --\\
			Cycle-IR~\cite{tan2019cycle} & \underline{0.4569} & --  & --  & -- & --\\
			GPDM~\cite{elnekave2022generating} & 0.7951 & 0.7043  & 0.4672  & 0.5188 & 0.6563\\
			Object-IR & \textbf{0.4266} & \textbf{0.3382} & \textbf{0.2975} & \textbf{0.4497} & \underline{0.6251} \\
			\bottomrule
		\end{tabular}%
		\label{tab:retargetme}%
	\end{table*}%

	We perform comparative experiments of the proposed Object-IR on our testing dataset and the RetargetMe benchmark~\cite{rubinstein2010comparative}. When performing our Object-IR on the RetargetMe dataset, we first downsample the input image to 224$\times$224 resolution and predict the mesh motion in the downsampled resolution. Then, we upsample the deformed mesh ($M_f$) and generate the retargeted image by warping the original input image using the upsampled mesh deformation.
	
	Many retargeting methods, e.g., seam carving (SC)~\cite{avidan2007seam}, Warp~\cite{wolf2007non}, SNS~\cite{wang2008optimized}, and Cycle-IR~\cite{tan2019cycle}, have published retargeting results (at different aspect ratios) on the RetargetMe benchmark. Thus, comparing our Object-IR with theirs in terms of visual quality is straightforward. For retargeting methods with available source codes, such as manual cropping (CR), simple scaling (SCL), SC~\cite{avidan2007seam}, SNS~\cite{wang2008optimized}, WSSCDNN~\cite{cho2017weakly}, and GPDM~\cite{elnekave2022generating}, we conduct comprehensive comparisons using both quantitative and visual quality assessment. 

	\begin{table*}
		\centering
		\caption{Quantitative comparisons on Satellite images.}
		\begin{tabular}{lccccc}
			\toprule
			Methods & 0.5$\times$  & 0.75$\times$ & 1.25$\times$ & 1.5$\times$  & 1.75$\times$ \\
			\midrule
			CR    & 0.4662 & \underline{0.2527} & --     & --    & -- \\
			SCL   & 0.5772 & 0.3814 & 0.3890 & 0.5622 & 0.7448 \\
			SC~\cite{avidan2007seam}    & 0.5028 & 0.3428 & \underline{0.3292} & \underline{0.4472} & 0.6110 \\
			SNS~\cite{wang2008optimized}    & \underline{0.4226} & 0.3308 & 0.3542 & 0.4574 & \underline{0.5920} \\
			WSSDCNN~\cite{cho2017weakly} & 0.4389 & 0.3557 & --    & --     & -- \\
			GPDM~\cite{elnekave2022generating}  & 0.5862 & 0.4795 & 0.3814 & 0.4954 & 0.6072 \\
			Object-IR & \textbf{0.4018} & \textbf{0.2115} & \textbf{0.2735} & \textbf{0.4131} & \textbf{0.5217} \\
			\bottomrule
		\end{tabular}%
		\label{tab:satellite}%
	\end{table*}%
	
	\subsection{Quantitative Comparison}

\begin{figure*}
		\centering
		\includegraphics[width=\linewidth]{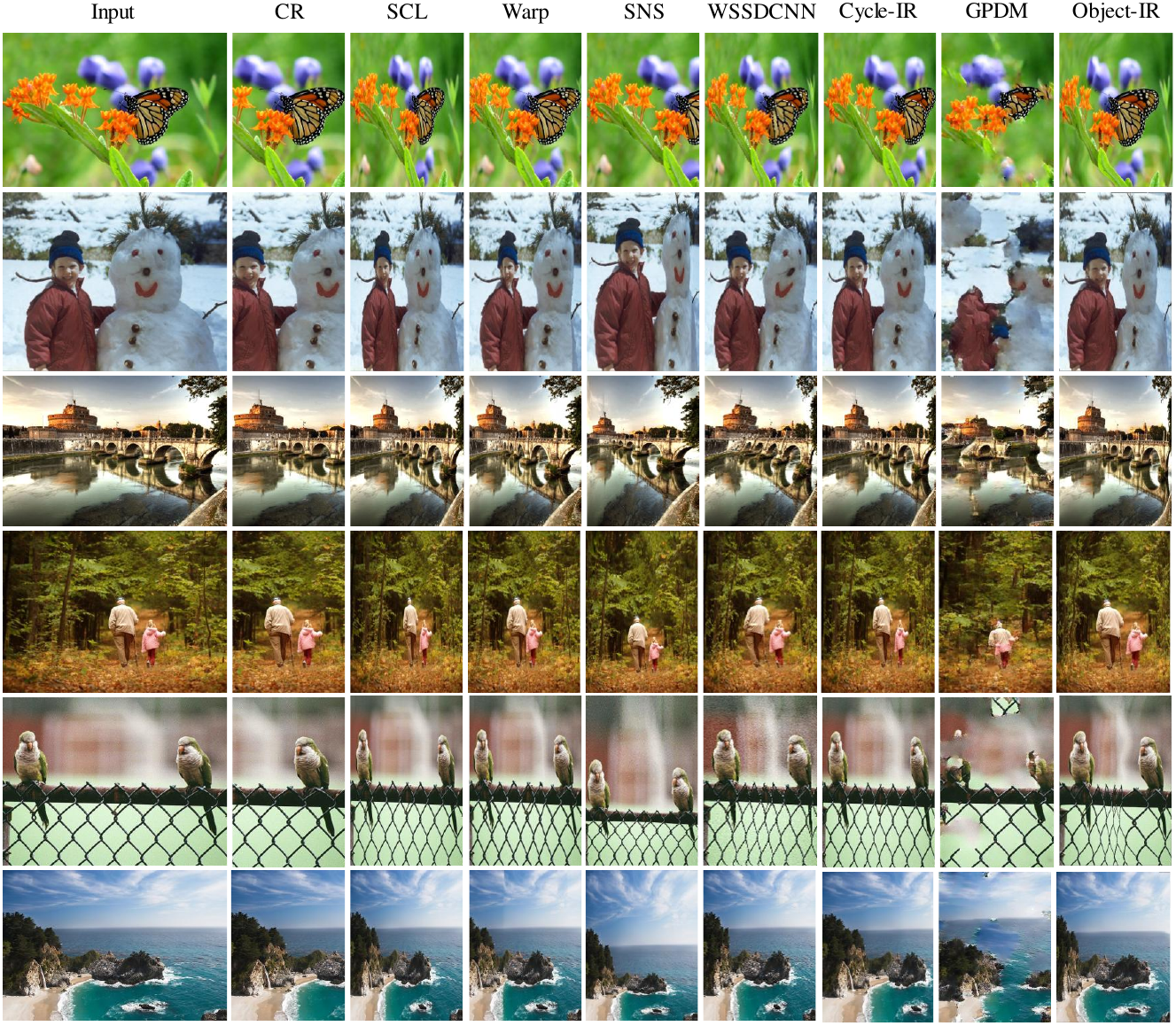}
		\caption{Visual comparisons of our Object-IR with representative retargeting methods. Images are retargeted to 0.5$\times$ width.}
		\label{fig:comp1}
	\end{figure*}

	To calculate the distortion errors of other retargeting methods, we detect objects in both the input and output and then design an object-matching algorithm, complemented by manual verification, for accurate evaluation. Images are resized to 0.5$\times$, 0.75$\times$, 1.25$\times$, 1.5$\times$, and 1.75$\times$ their original widths.  
	We compare quantitatively with other retargeting methods on our testing dataset and the RetargetMe dataset, as shown in Table \ref{tab:testing} and \ref{tab:retargetme}, where ``--'' indicates that no retargeted result can be generated by the method. 
	The best and second-best results are marked in \textbf{bold} and \underline{underlined}, respectively. The simple scaling (SCL) clearly produces the worst results since it lacks any content-aware design. GPDM also fails to mitigate the distortions in retargeted results due to object destruction issues, as shown in Fig. \ref{fig:comp1} and \ref{fig:comp2}.

	In Table \ref{tab:testing}, manual cropping (CR) yields the lowest distortion error for 0.75$\times$ width resizing. This is attributable to the concentrated distribution of small- to medium-sized objects within the input images of our testing dataset. However, CR may crop objects when resizing to 0.5$\times$ width, leading to higher distortion error. Compared to other methods, our Object-IR consistently attains the lowest distortion errors in image reduction and enlargement scenarios. As a result, it demonstrates the best retargeting quality. 
	
	In Table \ref{tab:retargetme}, we further evaluate our method on the RetargetMe dataset, whose images exhibit substantial divergence from those in the COCO dataset. Since Cycle-IR~\cite{tan2019cycle} published their results with 0.5$\times$ width resizing on RetargetMe, we evaluate the distortion error and report it as well. Manual cropping (CR) performs even worse than simple scaling (SCL) due to the presence of large, scattered objects in the images of RetargetMe. Other content-aware retargeting methods can mitigate distortions to some extent. Among them, our Object-IR still yields the lowest errors and thus has the best retargeting quality.
	
	We also evaluate the generalization ability to unseen domains, such as medical or satellite images. For medical images, it is difficult to define where the important objects are; thus, our distortion error metric is not suitable for quality evaluation of medical images. We only evaluate the retargeting quality on a satellite image dataset~\cite{yang2010bag}, the comparison results are shown in Table \ref{tab:satellite}. Our method still achieves the best retargeting quality, which shows robust cross-dataset generalization ability.

    	\begin{table*}
	\centering
	\caption{Ablation study of our Object-IR. The ``gray'' row corresponds to our Object-IR with default parameter settings.}
	\resizebox{\linewidth}{!}{
		\begin{tabular}{clcccccccc}
			\toprule
			\multirow{2}[4]{*}{ID} & \multirow{2}[4]{*}{Scale $\mathbf{s}$} & \multicolumn{3}{c}{Loss function} & \multicolumn{3}{c}{Mesh resolution} & \multicolumn{2}{c}{Dataset} \\
			\cmidrule{3-10}          &       & object loss $l_o$ & geometric loss $l_g$ & boundary loss $l_b$ & 4$\times$4   & 8$\times$8   & 16$\times$16 & Testing & \multicolumn{1}{c}{RetargetMe} \\
			\midrule
			1     &  --     & \checkmark     &       &       &       & \checkmark     &       & \underline{0.3384} & 0.5087 \\
			2     & $\mathbf{s}$=$\sqrt{0.5}$ & \checkmark     & \checkmark     &       &       & \checkmark     &       & \textbf{0.3310} & 0.4757 \\
			\rowcolor[gray]{0.8}
			3     & $\mathbf{s}$=$\sqrt{0.5}$ & \checkmark     & \checkmark     & \checkmark     &       & \checkmark     &       & 0.3471 & \textbf{0.4266} \\
			4     & $\mathbf{s}$=$\sqrt{0.5}$ & \checkmark     & \checkmark     & \checkmark     & \checkmark     &       &       & 0.4247 & 0.4877 \\
			5     & $\mathbf{s}$=$\sqrt{0.5}$ & \checkmark     & \checkmark     & \checkmark     &       &       & \checkmark     & 0.3875 & 0.5080 \\
			\midrule
			6     & $\mathbf{s}$=1     & \checkmark     & \checkmark     & \checkmark     &       & \checkmark     &       & 0.4165 & 0.5308 \\
			7     & $\mathbf{s}$=0.9  & \checkmark     & \checkmark     & \checkmark     &       & \checkmark     &       & 0.4001 & 0.4979 \\
			8     & $\mathbf{s}$=0.8   & \checkmark     & \checkmark     & \checkmark     &       & \checkmark     &       & 0.3659 & 0.4425 \\
			9     & $\mathbf{s}$=0.7   & \checkmark     & \checkmark     & \checkmark     &       & \checkmark     &       & 0.3636 & \underline{0.4269} \\
			10    & $\mathbf{s}$=0.6   & \checkmark     & \checkmark     & \checkmark     &       & \checkmark     &       & 0.3981 & 0.4458 \\
			11    & $\mathbf{s}$=0.5   & \checkmark     & \checkmark     & \checkmark     &       & \checkmark     &       & 0.4485 & 0.4891 \\
			\bottomrule
	\end{tabular}}
	\label{tab:ablation}%
\end{table*}
	
	\subsection{Visual Comparison}
	
\begin{figure*}
		\centering
		\includegraphics[width=\linewidth]{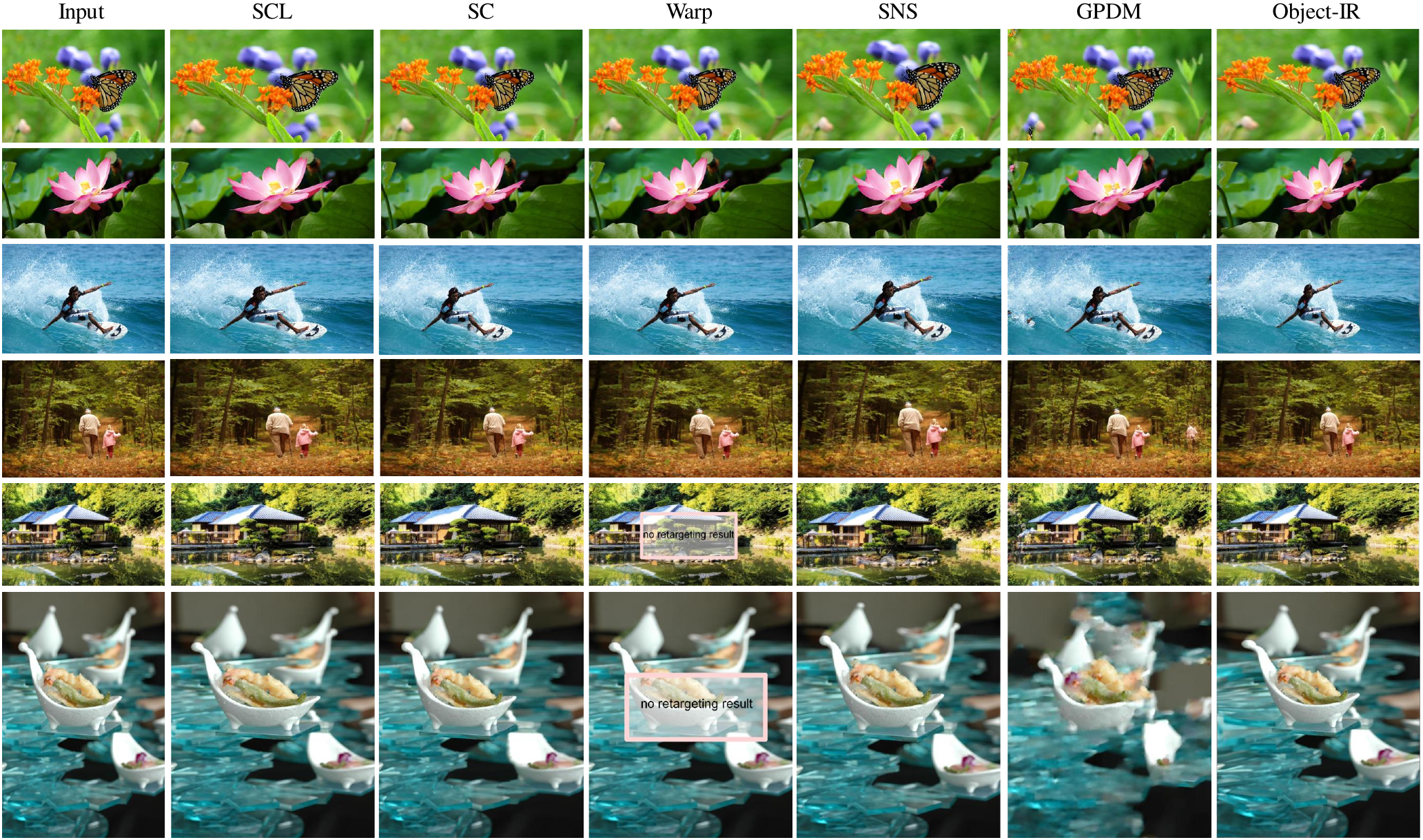}
		\caption{Visual comparison of our Object-IR with representative retargeting methods. Images are retargeted to 1.25$\times$ width. Note that for some images in RetargetMe, the warp method~\cite{wolf2007non} provides no result for 1.25$\times$ width resizing. We simply add an annotation to these images.}
		\label{fig:comp2}
	\end{figure*}

	We comprehensively compare the visual results with the other retargeting methods. Fig. \ref{fig:comp1} and \ref{fig:comp2} show several comparison results on the RetargetMe dataset. Manual cropping (CR) directly removes the content outside the objects in input images, leading to the loss of vital information. The SCL just merges adjacent pixels, resulting in severe distortions. The seam-carving method (SC)~\cite{avidan2007seam} may deform objects when seams are carved across them. The non-homogeneous warping method (Warp)~\cite{wolf2007non}, which is designed for video retargeting, has limited retargeting quality for images. The scale-and-stretch method (SNS)~\cite{wang2008optimized} manifests a certain degree of geometric preservation, which is consistent with the evaluation values in Table \ref{tab:retargetme}. However, it may alter the relative distribution of image contents, which differs substantially from other methods. GPDM~\cite{elnekave2022generating} generates retargeting results by minimizing patch distributions between input and output images. It may introduce fidelity inconsistencies, resulting in visually implausible results. Consequently, according to our quality metric, GPDM exhibits the worst distortions. Benefiting from the powerful representational ability of deep learning, WSSDCNN~\cite{cho2017weakly} and Cycle-IR~\cite{tan2019cycle} improved the retargeting quality further. In contrast, by exploiting object consistency for self-supervision and the high-deformation freedom of mesh deformation, our Object-IR yields the highest-quality retargeted results. Fig. \ref{fig:comp2} shows the comparison of the enlargement results. Input images are retargeted to 1.25$\times$ width. The comparison demonstrates that our Object-IR can still generate satisfactory results for image enlargement.
	
\begin{figure*}
		\centering
		\includegraphics[width=\linewidth]{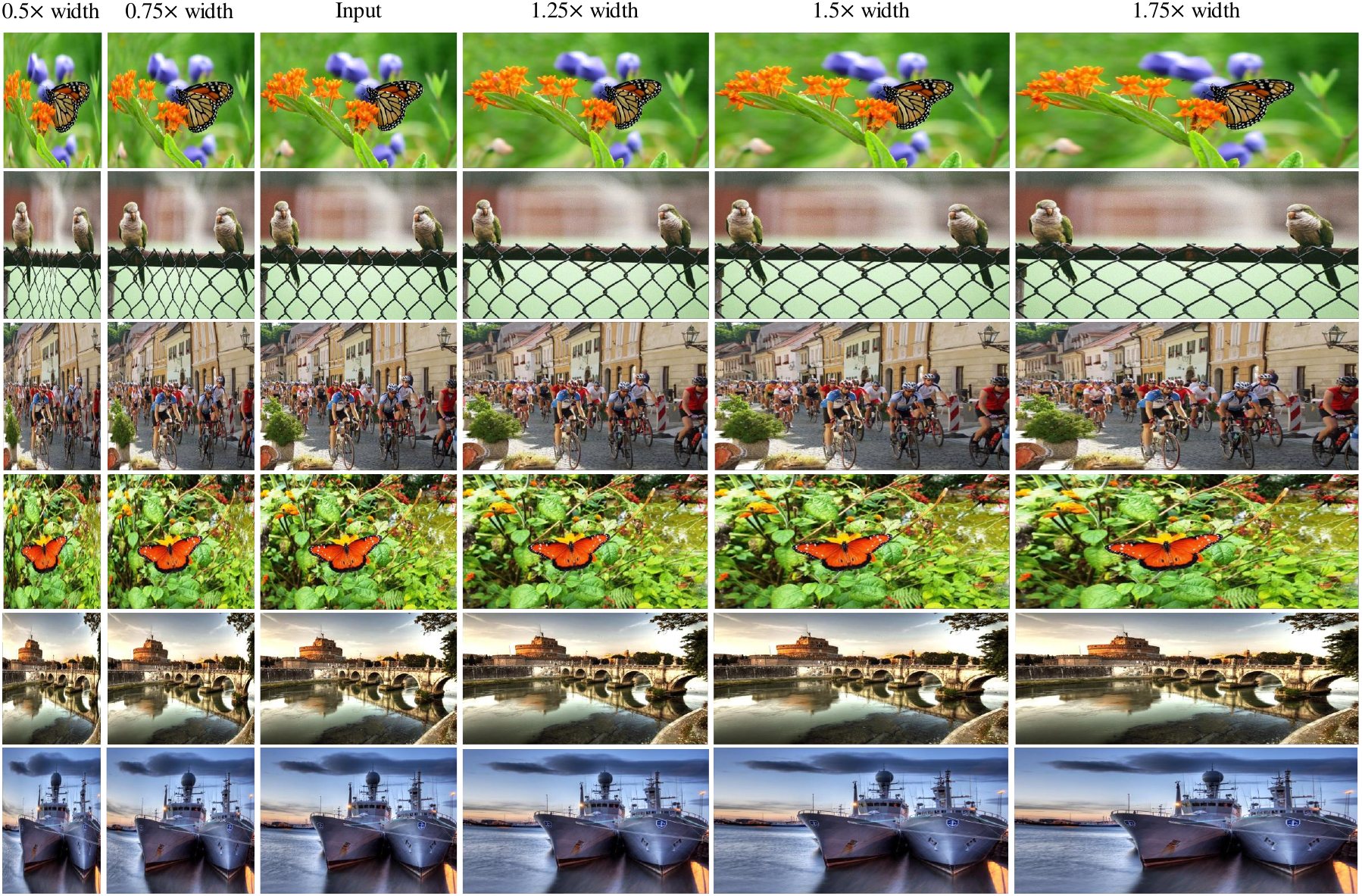}
		\caption{Visual examples of our Object-IR for arbitrary retargeting sizes. Even with a wide scale range (from 0.5 to 1.75), our Object-IR consistently generates high-quality retargeted images.}
		\label{fig:zoom}
	\end{figure*}

	Our Object-IR can handle input images of arbitrary sizes and retargeted images at arbitrary aspect ratios.
	Fig. \ref{fig:zoom} shows some visual examples where the width resizing scales are 0.5, 0.75, 1.25, 1.5, and 1.75, respectively. Despite the wide scale range, our Object-IR effectively preserves important areas and objects from distortion. These visual examples highlight the robust arbitrary-size retargeting capability of our Object-IR.

	\subsection{Ablation Study}
	
	We validate the effectiveness of every module in our Object-IR, as shown in Table \ref{tab:ablation}. Images in our testing and the RetargetMe datasets are resized to 0.5$\times$ width. Thus, the scale $\mathbf{s}$ is set to $\sqrt{0.5}$ by default.
	
	\begin{figure*}
		\centering
		\includegraphics[width=\linewidth]{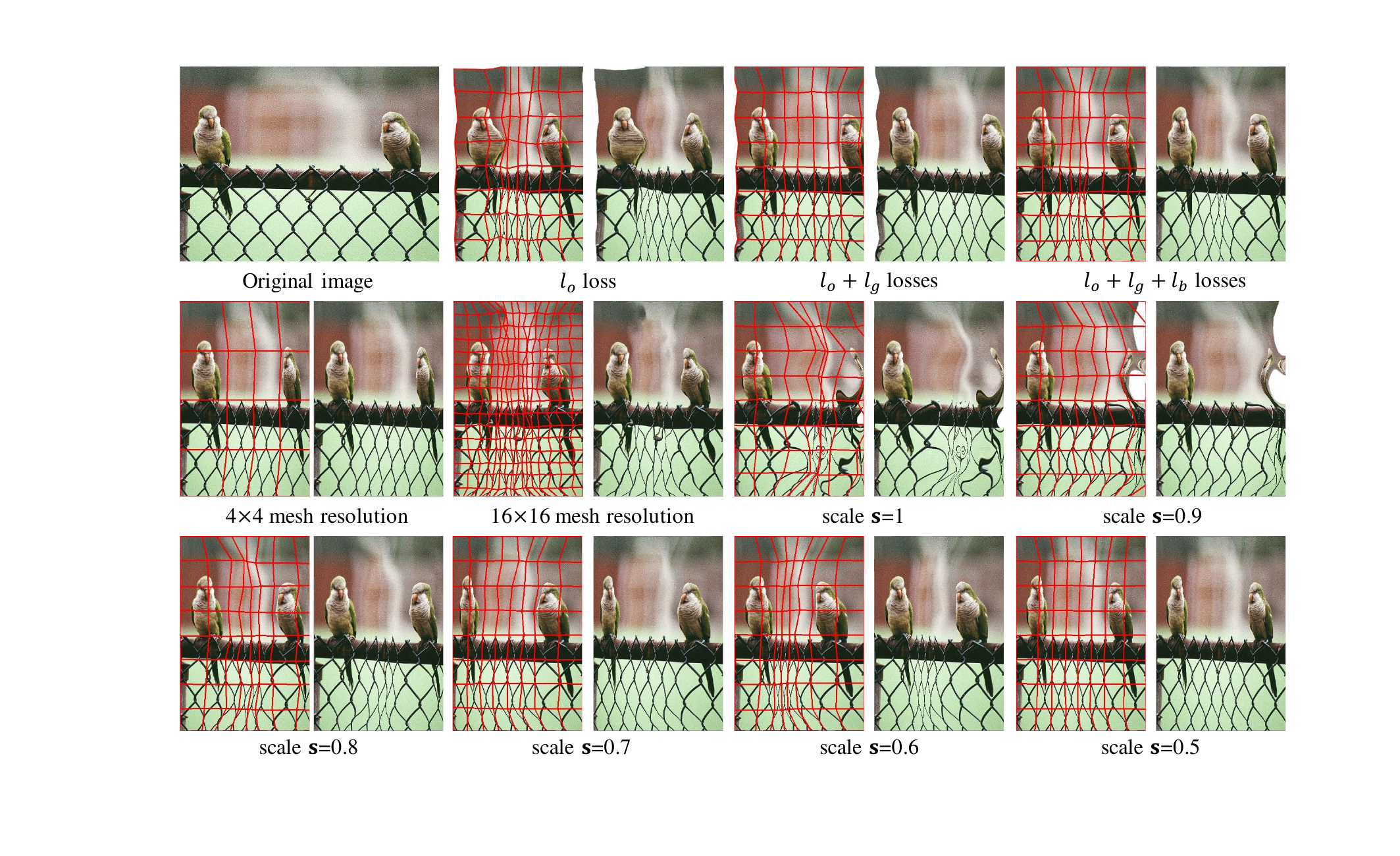}
		\caption{Visual comparisons of the ablation study. ``$l_o+l_g+l_b$ losses'' corresponds to our Object-IR with default parameter setting.}
	\label{fig:ablation}
\end{figure*}

\subsubsection{Objective function}

We ablate the geometric and boundary losses as the basic structure and evaluate the effectiveness of different losses in the objective function. As demonstrated in experiments 1-3 of Table \ref{tab:ablation}, incorporating the geometric loss improves the retargeting quality for both our testing and the RetargetMe datasets. In our testing dataset, adding the boundary loss has a somewhat opposite impact on geometry preservation based on the metric. However, it effectively enforces a visually appealing rectangular output. Fig. \ref{fig:ablation} (first row) shows a visual comparison of different constraint losses, where the deformed meshes are also presented.

\subsubsection{Mesh resolution}

We assess the retargeting quality of our Object-IR at mesh resolutions of 4$\times$4, 8$\times$8, and 16$\times$16, as shown in experiments 3-5 of Table \ref{tab:ablation}. Reducing mesh resolution restricts the deformation capacity of mesh vertices during image resizing. Conversely, increasing mesh resolution often requires additional constraints to prevent distortion of the numerous meshes. In our experiment, adopting 8$\times$8 mesh resolution achieves a good balance between computational efficiency and retargeting performance, as shown in the visual example in Fig. \ref{fig:ablation}.

\subsubsection{Scale hyper-parameter}

We further evaluate the performance of our Object-IR under diverse settings for hyper-parameter $\mathbf{s}$, as shown in experiments 3 and 6-11 of Table \ref{tab:ablation}.  When $\mathbf{s}$=1, objects in the retargeted results are constrained to maintain full scale, which is unfeasible for 0.5$\times$ width resizing. Consequently, this leads to significant distortions, as exemplified in Fig. \ref{fig:ablation}. Decreasing $\mathbf{s}$ can improve the performance, yet continuously reducing the parameter $\mathbf{s}$ may yield the opposite result. The experiments verify that, for model training at arbitrary retargeting sizes, adaptive parameter settings (e.g., Eq. (\ref{eq:s})) outperform fixed values in terms of effectiveness and robustness.

\subsection{User Study}

\begin{table*}
	\centering
	\caption{User study for comparing Object-IR with other representative retargeting methods.}
	\resizebox{\linewidth}{!}{
		\begin{tabular}{lcccccccccccc|c|c}
			\toprule
			Methods & P1    & P2    & P3    & P4    & P5    & P6    & P7    & P8    & P9    & P10   & P11   & P12   & Total & Prefer \\
			\midrule
			SCL   & 16    & 15    & 23    & 19    & 13    & 24    & 16    & 20    & 18    & 26    & 22    & 25    & 237   & 39.50\% \\
			SC~\cite{avidan2007seam}   & 22    & 21    & 20    & 14    & 20    & 24    & 20    & 17    & 20    & 15    & 12    & 14    & 219   & 36.50\% \\
			Warp~\cite{wolf2007non}  & 22    & 23    & 22    & 22    & 28    & 23    & 24    & 23    & 24    & 23    & 20    & 22    & 276   & 46.00\% \\    
			SNS~\cite{wang2008optimized}  & 23    & 22    & 28    & 25    & 25    & 23    & 23    & 25    & 26    & 22    & 28    & 26    & 296   & 49.33\% \\
			Cycle-IR~\cite{tan2019cycle} & 31    & 34    & 23    & 32    & 30    & 26    & 33    & 27    & 30    & 25    & 30    & 30    & 351   & \underline{58.50\%} \\
			Object-IR  & 36 & 35 & 34 & 38 & 34 & 30 & 34 & 38 & 32 & 39 & 38 & 33 & 421 & \textbf{70.17\%} \\
			\bottomrule
	\end{tabular}}
	\label{tab:user}%
\end{table*}%

To validate the superiority of our Object-IR and ensure that the proposed retargeting quality assessment aligns with human vision, we conduct a user study to evaluate whether the users prefer our results. We choose 6 representative retargeting methods including SCL, SC~\cite{avidan2007seam}, SNS~\cite{wang2008optimized}, Warp~\cite{wolf2007non}, Cycle-IR~\cite{tan2019cycle}, and our Object-IR. 
We invite 12 participants to evaluate the retargeting performance of each method in terms of visual quality (geometric distortions and artifacts). 
To mitigate the evaluation burden on participants, we randomly select 10 images from the RetargetMe dataset. These images are resized to 0.5$\times$ width using the 6 methods. Other resizing scales like 0.75 and 1.25 are excluded from the user study because our prior analysis indicates that visual discrimination of differences at these scales is challenging. 
During the study, one original image and two retargeted images produced by two out of six methods are randomly shown to the participants. Afterward, the participants select one retargeted image they prefer.

The study requires $\mathrm{C}_6^2\times 10\times12=1,800$ comparisons in total. Each participant is required to compare $\mathrm{C}_6^2\times 10=150$ times. Any two methods are compared $10\times 12=120$ times. Each method has $5\times 10 \times 12=600$ comparisons. Table \ref{tab:user} reports the statistical results of the user study. Each value in the table represents the number of times a method is preferred by the participant.  The study indicates that our Object-IR received 421 votes out of 600 in all comparisons, accounting for 70.17\% (421/600), ranking highest among the methods. In comparison, the SCL, SC, SNS, Warp, and Cycle-IR methods have preference rates of 39.50\%, 36.50\%, 49.33\%, 46.00\%, and 58.50\%, respectively. The results of the user study are consistent with the quantitative and visual evaluations. They further validate that our retargeting quality assessment is in line with human visual perception.

\subsection{Evaluation for Retargeting Quality Assessment}

Our newly proposed distortion metric is intuitive and closely aligned with object bounding box ratios. To theoretically analyze its effectiveness, we conduct a Pearson correlation analysis comparing the metric with user study scores. As there are currently very few image quality assessment (IQA) metrics specifically designed for image retargeting, we also employed several widely used full-reference and no-reference IQA metrics, including BRISQUE~\cite{mittal2012no}, NIQE~\cite{mittal2012making}, PIQE~\cite{venkatanath2015blind}, HyperIQA~\cite{su2020blindly}, and CMMD~\cite{jayasumana2024rethinking}, to evaluate the retargeted results considered in the user study. The results, presented in Table \ref{tab:pearson}, show that our method consistently achieves the highest quality across all metrics. Moreover, our proposed metric demonstrates a strong linear correlation with user study scores, underscoring its reliability as a proxy for human perceptual judgment.

\begin{table*}
	\centering
	\caption{Retargeting quality assessment via different IQA metrics and Pearson correlation study.}
	\begin{threeparttable}
		\resizebox{\linewidth}{!}{
			\begin{tabular}{lccccccc}
				\toprule
				& User study score $\uparrow$ & BRISQUE $\downarrow$ & NIQE $\downarrow$ & PIQE $\downarrow$ & HyperIQA $\uparrow$ & CMMD $\downarrow$ & Our metric $\downarrow$ \\
				\midrule
				SCL   & 39.50  & 18.99 & 4.3543 & 40.52 & 60.13 & 0.322  & 0.5930  \\
				SC~\cite{avidan2007seam}    & 36.50  & 23.72  & 4.5528 & 45.28 & 54.16 & 0.406 & 0.6448 \\
				Warp~\cite{wolf2007non}  & 46.00  & 19.80  & 4.3225 & 41.86  & 65.17 & 0.328 & 0.4336 \\
				SNS~\cite{wang2008optimized}   & 49.33 & 19.28 & 4.3019 & 40.28 & 66.70  & 0.314 & 0.4275 \\
				Cycle-IR ~\cite{tan2019cycle} & 58.50  & 17.46 & 4.2770  & 36.15 & 67.21 & 0.289 & 0.4011 \\
				Object-IR & \textbf{70.17} & \textbf{15.12} & \textbf{3.8323} & \textbf{33.82} & \textbf{74.88} & \textbf{0.252} & \textbf{0.3259} \\
				\midrule
				Pearson coefficient\tnote{1} &  --  & 0.8875 & 0.9186  & 0.9397 & 0.9394 & 0.8865 & 0.9107 \\
				p-value & -- & 0.0183  & 0.0097 & 0.0053 & 0.0054 & 0.0186 & 0.0116  \\
				\bottomrule
		\end{tabular}}
		\begin{tablenotes}
			\footnotesize
			\item[1] Note that all the Pearson coefficients have been reversed to positive values for clear comparison.
		\end{tablenotes}
	\end{threeparttable}
	\label{tab:pearson}%
\end{table*}

\subsection{Time Efficiency}

We compare the computational efficiency of our Object-IR with other representative retargeting methods. Experiments are tested with an Intel i9-11900K 3.5GHz CPU and NVIDIA RTX 3090 GPU.
Table \ref{tab:time} presents the results. The tests are conducted on the RetargetMe dataset for 0.5$\times$, 0.75$\times$, 1.25$\times$, 1.5$\times$, and 1.75$\times$ width resizing. The seam-carving (SC) method~\cite{avidan2007seam} is the most time-consuming. It iteratively removes or adds unnoticeable seams in the input images. Additionally, the elapsed time increases as the width is resized to a larger or smaller value. Similarly, GPDM~\cite{elnekave2022generating} suffers from a time-consuming problem due to the extensive patch matching required between the input and output images. The traditional warping-based method, SNS~\cite{wang2008optimized}, requires significantly less time than SC and GPDM. 
In contrast, learning-based methods, WSSDCNN~\cite{cho2017weakly} and our Object-IR, can be accelerated by a GPU, achieving speeds far surpassing traditional methods, which a GPU cannot accelerate. In summary, our Object-IR takes the least time and exhibits the least variation across different width resizing scenarios.

\begin{table*}
	\centering
	\caption{Comparison of elapsed time among different retargeting methods (seconds).}
	\begin{tabular}{lccccc}
		\toprule
		Methods & 0.5$\times$   & 0.75$\times$  & 1.25$\times$  & 1.5$\times$ & 1.75$\times$\\
		\midrule
		SC~\cite{avidan2007seam}   & 154.409  & 90.634 & 185.931 & 328.260  &  421.654 \\
		SNS~\cite{wang2008optimized}  & 4.057 & 2.364 & 2.350 & 4.010 & 4.256\\
		WSSDCNN~\cite{cho2017weakly} & \underline{0.300} & \underline{0.286} & -- & -- & --\\
		GPDM~\cite{elnekave2022generating} & 25.115  & 31.017 & 32.202 & 35.776 & 69.262\\
		Object-IR & \textbf{0.025} & \textbf{0.027} & \textbf{0.032} & \textbf{0.033} & \textbf{0.035}\\
		\bottomrule
	\end{tabular}
	\label{tab:time}%
\end{table*}

\subsection{Limitation and Discussion}

Similar to all retargeting methods, the results may exhibit distortions due to the prevalence of salient objects or widespread geometric structures in the image.
	We manually inspected all retargeted results on the RetargetMe benchmark and evaluated the failure rate. A result is considered a failure if clear distortions are observed. We then compared the failure rates across different retargeting methods, and the results are reported in Table \ref{tab:failure}. Our Object-IR consistently achieves the lowest failure rate across various retargeting ratios.

\begin{table*}
	\centering
	\caption{Failure rate analysis for retargeted results on the RetargetMe benchmark.}
	\begin{tabular}{lccccc}
		\toprule
		Methods & 0.5$\times$  & 0.75$\times$ & 1.25$\times$ & 1.5$\times$  & 1.75$\times$ \\
		\midrule
		SC~\cite{avidan2007seam}   & 87.50\% & 58.75\% & 65.00\% & 86.25\% & 91.25\% \\
		SNS~\cite{wang2008optimized}    & 53.75\% & 27.50\% & 25.00\% & 56.25\% & 63.75\% \\
		WSSDCNN~\cite{cho2017weakly} & 77.50\% & 36.25\% & --    & --    & -- \\
		GPDM~\cite{elnekave2022generating}  & 97.50\% & 91.25\% & 96.25\% & 97.50\% & 98.75\% \\
		Cycle-IR~\cite{tan2019cycle} & 46.25\% & --    & --    & --    & -- \\
		Object-IR  & \textbf{32.50\%} & \textbf{25.00\%} & \textbf{22.50\%} & \textbf{37.50\%} & \textbf{48.75\%} \\
		\bottomrule
	\end{tabular}%
	\label{tab:failure}%
\end{table*}%

\begin{figure*}
	\centering
	\includegraphics[width=\linewidth]{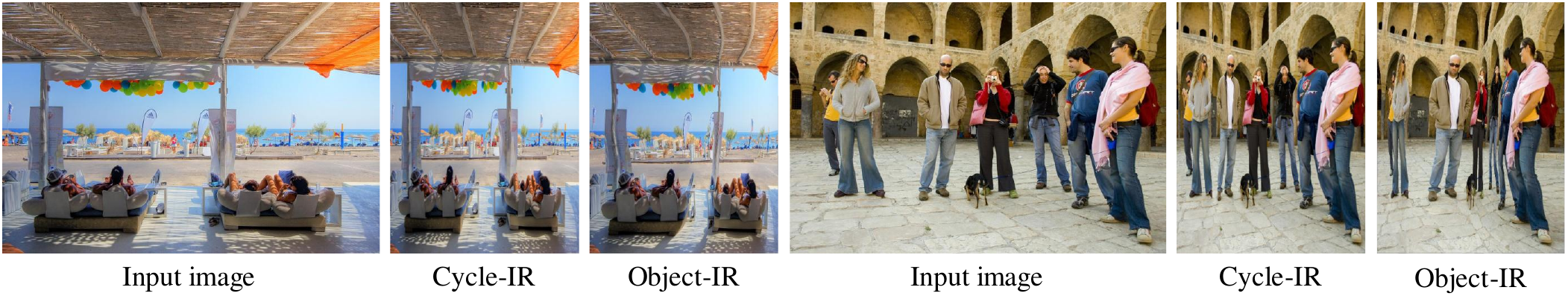}
	\caption{Failure cases of our Object-IR. Images are resized to 0.5$\times$ width.}
	\label{fig:failure}
\end{figure*}

Fig. \ref{fig:failure} shows two failure cases of our Object-IR. In the left-hand case of Fig. \ref{fig:failure}, semantically important objects and line structures are overly scattered. Our method fails to fully preserve all the structures from being distorted, while Cycle-IR may produce a more visually pleasing result. The right-hand case in Fig. \ref{fig:failure} represents another failure scenario, where even a minor distortion of humans is readily noticeable to observers. Our Object-IR can only effectively prevent one person in the middle from being distorted. Moreover, though our Object-IR demonstrates the best retargeting qualities in Tables \ref{tab:testing}, \ref{tab:retargetme}, and \ref{tab:satellite}. The cross-dataset generalization ability and robustness can be further enhanced.

We identify three complementary directions for further improving the retargeting quality of our Object-IR.
	
First, on the architectural side, our current method employs a simple ResNet-50 for feature extraction and a fully connected layer for motion regression, but more expressive designs such as transfer learning~\cite{sakirin2025application}, transformers~\cite{cao2023recurrent}, or graph convolutional networks (GCNs)~\cite{brody2022attentive} could be explored. Attention mechanisms and GCNs provide natural ways to model spatial or topological relationships among mesh grids, which may lead to more faithful retargeting. In addition, our method currently adopts an $8 \times 8$ mesh resolution to balance efficiency and performance. Employing dynamic mesh resolution or predicting optical flow for pixel-wise deformation could further benefit scenes with dense object layouts.
	
Second, on the objective side, incorporating richer semantic and geometric constraints could provide stronger supervision during training. Potential directions include perceptual feature consistency using VGG (as in Cycle-IR~\cite{tan2019cycle}), CLIP-based feature extraction~\cite{li2024clip,Yang_2025_CVPR}, and explicit preservation of line structures to protect critical regions from distortion.
	
Third, on the evaluation side, we note that existing IQA metrics are not tailored for image retargeting and often fail to capture retargeting-specific distortions. The proposed distortion error is limited in its capacity to measure the aspect ratio of objects; further consideration should be given to incorporating measures of geometric distortions at finer scales. Designing a dedicated metric, analogous in spirit to recent advances in aesthetic assessment~\cite{zhang2024confidence}, would enable more accurate and perceptually aligned evaluation. Such a metric could incorporate both semantic preservation and geometric consistency, providing stronger guidance for model development and fairer comparisons across methods.
	
We view these architectural, objective-level, and evaluation-oriented enhancements as promising directions for future research and plan to investigate them in subsequent work.

\section{Conclusion}
\label{sec:conclude}

In this paper, we present a novel image retargeting approach named Object-IR. This method integrates object consistency and mesh deformation within a self-supervised learning framework. We formulate image retargeting as a learnable mesh-based warping, where the neural network directly estimates the deformed mesh from the rigid mesh defined in the output resolution. For model training, we propose a comprehensive objective function comprising object consistency, geometric preservation, and rectangular output enforcement. Additionally, we introduce a retargeting quality assessment to evaluate distortion errors in the retargeted results. Extensive experimental comparisons on our dataset and the RetargetMe benchmark, including quantitative and qualitative analyses along with a user study and speed comparison, validate the superiority and robustness of our Object-IR relative to other state-of-the-art retargeting methods. Finally, we discuss the limitations and propose several directions to further improve our Object-IR in the future.

\appendix
%

\section*{Acknowledgments}
This work is partially supported by the Natural Science Foundation of Henan Province under Grant 222300420140 and the Institute for Complexity Science, Henan University of Technology, under No. CSKFJJ-2025-10.

\section*{CRediT authorship contribution statement}

\textbf{Tianli Liao}: Conceptualization, Methodology, Supervision. \textbf{Ran Wang}: Data curation, Software, Writing-Original draft preparation. \textbf{Siqing Zhang}: Data curation, Software, Writing-Reviewing and Editing. \textbf{Lei Li}: Visualization, Writing-Reviewing and Editing. \textbf{Guangen Liu}: Software, Validation, Writing-Reviewing and Editing. \textbf{Chenyang Zhao}: Software, Writing-Reviewing and Editing. \textbf{Heling Cao}: Software, Writing-Reviewing and Editing. \textbf{Peng Li}: Supervision, Writing-Reviewing and Editing.


\bibliographystyle{elsarticle-num} 
\bibliography{refs}



%
%
%
\end{document}